\documentclass{article}
\usepackage{spconf,amsmath,graphicx,url}
\RequirePackage{fix-cm}
\usepackage{graphicx}
\usepackage{algorithm}
\usepackage[noend]{algpseudocode}
\makeatletter
\def\BState{\State\hskip-\ALG@thistlm}
\makeatother
\usepackage{color}

\title{Predicting Future Pedestrian Motion in Video Sequences Using Crowd Simulation - vai para Frontiers}
\name{Cliceres Mack Dal Bianco  and Soraia Raupp Musse\thanks{Thanks to Brazilian agencies: CAPES, CNPQ and FAPERGS.}} 
\address{VHLab -- Computer Science Department, \\
		Pontifical Catholic University of Rio Grande do Sul, Brazil\\
 }

\begin{document}

\maketitle

\begin{abstract}
While human and group analysis have become an important area in last decades, some current and relevant applications involve to estimate future motion of pedestrians in real video sequences. This paper presents a method to provide motion estimation of real pedestrians in next seconds, using crowd simulation. Our method is based on Physics and heuristics and use BioCrowds as crowd simulation methodology to estimate future positions of people in video sequences. Results show that our method for estimation works well even for complex videos where events can happen. The maximum achieved average error is $2.72$cm when estimating the future motion of 32 pedestrians with more than 2 seconds in advance. This paper discusses this and other results.
\end{abstract}


\section{Introduction}

Crowd analysis is a phenomenon of great interest in a large number of applications. Surveillance, entertainment and social sciences are examples of fields that can benefit from the development of this area of study. Literature dealt with different applications of crowd analysis, for example counting people in crowds~\cite{Chan2009} and~\cite{cai2014}, group and crowd movement and formation~\cite{Zhou2014},~\cite{Ricky:15} and~\cite{jo2013review} and detection of social groups in crowds~\cite{Shao2014}, ~\cite{Feng2015} and~\cite{Chandran2015}. 

A nowadays interesting problem is the possibility of estimating future crowd and group motion in a certain scenario given its current state. It can be useful to estimate future positions where robots and surveillance cameras should point out and try to predict possible problems with application in safety systems. Some researchers are interested about to estimate pedestrian motion, where some of them take benefit of external equipments to help on that. For instance, a recent work proposed by Khomchuk and collaborators~\cite{Khomchuk2016} presents a technique for estimating pedestrians direction of motion using MD (Micro-Doppler) radar signatures. In addition, the authors use a supervised regression to estimate the mapping between the directions of motion and the corresponding MD signatures. In a previous work Bianco et al.~\cite{Bianco2016} propose a technique to fast forward the behaviors of crowds, in the context of numerical simulation, providing a future estimation of virtual humans motion. Indeed, crowd simulation has been investigated in many applications over the last years. 
Despite all interest in the entertainment area, crowd simulation tools are also important in security applications. This relevant area aims to investigate the impact of a high number of people behaving in a specific environment to improve people security and comfort.

In this work we propose an extension to the method proposed by Bianco et al~\cite{Bianco2016}, where the authors present a model for estimating next positions of virtual humans simulated with crowd methods.
The main question we want to discuss in this paper is concerning with the prediction of the future crowd behavior in real video sequences, but dealing with not well behaved situation, i.e. when people flow features are not kept until the end of sequence, probably because some events happen and impact the expected pedestrian motion. 

This paper presents an extension to~\cite{Bianco2016} in order to provide the estimative for future pedestrian motion in video sequences dealing with the possibility that events can happen and behavioral patterns can change. The main applications of this work are: investigations on security systems dealing with the possibility to know in advance crowd issues that can emerge, e.g. high densities; and possible applications in movie and animations as well, since our method can also be used as a tool to movie continuation, finding out parameters that allow to provide a synthetic sequence of a movie.
The remaining of this paper presents some related work (Section ~\ref{sec:related}),  while 
Section~\ref{sec:modelEvents} describes our model to estimate positions considering events. 
Finally, Sections ~\ref{sec:results} and ~\ref{sec:final} present results and final considerations.

\section{Related Work}
\label{sec:related}

Some works have focused on the prediction of future behavior, as in~\cite{Yi:2015}, where the authors estimate the time taken by people to achieve their goals. For this, some characteristics need to be identified, such as regions of interest, origin and destination, people blocking the flow, as well as distribution of global density and speeds. Afterwards, using  statistical calculations, the authors estimate the time to reach the goal. As said before, Khomchuk and collaborators~\cite{Khomchuk2016} use radar signatures (MD) through a supervised regression to estimate the mapping between the directions of motion and the corresponding MD signatures. Keller and Gavrila~\cite{Keller2014} present a study on pedestrian path prediction at short subsecond time intervals. Their approaches are based on Gaussian process dynamical models and probabilistic hierarchical trajectory matching using a Kalman filter. Crowd simulation is the process by which the movement of many agents are calculated for the purpose of, for example, filling and animating virtual scenes ~\cite{DalBianco:2017:PMF:3044431.2996202,Thalmann2013} or verifying the security of real environments~\cite{Almeida:7524675}. 

In Bianco et al.~\cite{Bianco2016}, the authors estimate the individual position of virtual people based on a prior information of goals and speeds. In that work, prior positions come from a dead reckoning method and are later adjusted using a two-step method, in which the global environment complexity (EC) and the interaction with other people (IP) are considered. In this work we propose a method inspired on~\cite{Bianco2016}, but applied and tested in real video sequences. It offers some challenges as the fact that goals, speeds and environment complexity are not previous known as in crowd simulation. Next section presents the method described in~\cite{Bianco2016}.

\subsection{The FF Model for Estimation of Crowd Behavior}
\label{sec:model}

In this section we summarize the previous approach~\cite{Bianco2016} to provide  estimation for future crowd behavior, in simulation, as shown in  Algorithm~\ref{TM}. A Pedestrian Dead Reckoning method based on Physics
($PDR$ on line 5) is initially used to estimate future positions for agents in the crowd (using goals, positions and speeds). In addition to that term, crowd position estimation should also take into account the environment complexity ($EC$ - line 6), i.e. the free region and presence of obstacles. Also, agents can be affected by others. The $IP$ step (line 7 in  Algorithm~\ref{TM}) describes how individual velocities should be affected by the presence of other agents. In this latter case, the previous approach proposed the people impact on each other is based on Weibull distribution~\cite{Bianco2016}. Finally, the last step, called \textit{Repositioning} (line 8), is responsible for fine tuning the agents' positions in the environment, avoiding collisions among them and obstacles in the repositioning process. 

The basic data required to estimate future positions for agents in simulations are:
\begin{itemize}
\item Agent $i$ position at each frame $t$: $\vec{X}_{t,i}$;
\item the time $it$ of the beginning of the simulation;
\item the time $t>it$ when the simulation pauses and the Fast Forward behavior starts;
\item the time $t+\Delta t$ when the FF method should generate the final positions (repositioning) and then the simulation keeps going until the end. 
\end{itemize}
Please refer to~\cite{Bianco2016} for further details.

\begin{algorithm}
\footnotesize
\label{algo1}
\caption{FF Method}\label{TM}
\begin{algorithmic}[1]
\Procedure{ }{}
\State Continuous Simulation starts at frame $it$ and stops at frame $t$; 
\State FF should run to compute positions for all agents at frame $t+\Delta t$
\BState \emph{loop}: For each agent $i$ at frame $t+\Delta t$
\State $\vec{X}_{t+\Delta t,i}^1 \gets PDR(\vec{X}_{t,i})$;
\State $\vec{X}_{t+\Delta t, i}^2 \gets EC(\vec{X}_{t+\Delta t,i}^1)$;
\State $\vec{X}_{t+\Delta t,i}^3 \gets IP(\vec{X}_{t+\Delta t,i}^2)$;
\State Repositioning($\vec{X}_{t+\Delta t,i}^3$);
\State $i \gets i+1$.
\State \textbf{goto} \emph{loop}.
\State \textbf{close};
\State End.
\EndProcedure
\end{algorithmic}
\end{algorithm}

\section{Our Time Machine Method (TM)}
\label{sec:modelEvents}
In this paper, we developed a method that can easily and quickly estimate future pedestrian paths, even when events happen causing changes in expected crowd flow. For proof of concept, we used BioCrowds, as related in~\cite{Bianco2016}. However, the method could be integrated in any crowd simulation platform.

First of all, 
propose to evaluate the results using Equation~\ref{eq:Error1}, which provides the error in $centimeters$ 
during the period of the "travel in the time machine", i.e. if the video stops at frame 60 and we estimate at frame 110, Equation~\ref{eq:Error1} describes the implied error obtained when estimating the future position of individual $i$ in frame $t+\Delta t$, given that video stops at frame $t$: 
\begin{equation}
E_{t\rightarrow t+\Delta t,i} = d( \vec{TM}_{t+\Delta t,i}, \vec{X}_{t+\Delta t,i}),
\label{eq:Error1}
\end{equation}



where $d(\cdot,\cdot)$ is the Euclidean between two vectors, 
$\vec{TM}_{t+\Delta t,i}$ and $\vec{X}_{t+\Delta t,i}$ are respectively the positions of same  individual $i$ in frame $t+\Delta t$ (computed using the $TM$ method) and the video sequence, respectively. The obtained value $E_{t\rightarrow t+\Delta t,i}$ provides a way to compare our method with real video sequences. 
Error value is measured in $centimeters$ and the lower the value (minimum $0$), the better the result.

\subsection{First Step: Capturing Data from Video Sequences}
\label{sec:tracking}

In order to have information to estimate the future positions of individuals in a video sequence, we proceed with a tracking step, as described in~\cite{Favaretto:2016}. The people initial detection is performed using the work proposed by Viola and Jones~\cite{viola2001}. The boosted classifier working with haar-like features was trained with 4500 views of people heads as positive examples, and 1000 negative (CoffeBreak and Caviar Head datasets from Tosato et al.~\cite{dataset:coffe} were used). This detector performs the initial position detection of people based on their heads, which are the input parameters for the next step: tracking. Once the individuals are detected, the trajectories are obtained using a method proposed by Bins et al.~\cite{Bins2013}. Then, the primarily extracted information for each individual $i$, in the video, at each timestep $t$ is the 2D position $X_{t,i}$. In addition, we compute the individual motion vector $m_{t,i} = X_{t,i} - X_{t-\alpha,i}$, where $\alpha$ is related to the size of motion vector (we empirically defined $\alpha=5$) and $t$ is the last frame of filmed sequence to be used in the estimation process and serve as the initial frame for the positions estimation. Other two information are needed at this time: the average size of individuals in video $v$ (manually informed) and the total number of individuals present in the last filmed frame (captured given the tracking method).

Our method works for static cameras. The image is discretized into cells in a grid. The user defines the grid resolution and set (manually) the regions where there are obstacles (see Figure~\ref{fig:cena}(d)). This information is going to be used in Estimation step to provide the environment complexity, i.e. a relationship among the occupied area (with obstacles) and the total filmed area.

\subsection{Second Step: Estimating Future Positions}

In~\cite{Bianco2016} the virtual humans goals and speeds are known, so the method can use Physics to provide a prior position for each individual. In our case, we use the motion vector, acquired in the tracking phase, to use as prior, to estimate individuals positions at a desired estimated time $t_e$, as follows:

\begin{equation}
\vec{X}^1_{t_e,i} = \vec{X}_{t,i} + \vec{m}_{t,i} . (t_e - t),
\label{eq:PDR1}
\end{equation}
where $\vec{X}^1_{t_e,i}$ is the position of agent $i$ at predicted time $t_e$, only based on Physics. $\vec{X}_{t,i}$ is the position of agent $i$ in frame $t$ (last frame where positions were tracked in the video), $\vec{m}{t,i}$ is the agent motion vector, computed in the last captured frame. In addition, $t_e-t$ represents the time difference between last tracked frame $t$ and the desired estimated time $t_e$. Equation~\ref{eq:PDR1} presents our adaptation for line $5$ mentioned in Algorithm 1. 

The environment complexity is modeled taken into account the free space for crowd movement, i.e. considering number of agents and obstacles and total space: $EC_v = \min\{1,\frac{O_v}{T_v}\}$,
where $O_v$ represents the occupied percentage of total space (manually informed based on grid), filmed in the video sequence $v$, while $T_v$ represents the total filmed space, i.e. $=100$. If $EC_v >= 1$ then no motion is allowed. Equation 3 
presents our proposal to use $EC_v$ as a penalty function in position estimation of individual $i$, adapting the Equation~\ref{eq:PDR1} at estimated time $t_e$:
\begin{eqnarray}
\vec{X}^2_{t_e,i}&=& \vec{X}^1_{t,i} + \vec{m}_{t,i} . (1 - EC_v). (t_e - t).
\label{eq:ECPos}
\end{eqnarray}
In~\cite{Bianco2016}, authors found that the impact of other people in each individual motion resembles the Weibull distribution, given by $f(x) = \frac{b}{a}\left(\frac{x}{a}\right)^{b-1}\exp\left [{-\left(\frac{x}{a}\right)^b}\right ]$.
They used maximum likelihood estimation in order to find the values of $a$ and $b$ that best fit the Weibull distribution to the data. After this process, they achieved 10 Weibull 
$f(\Delta \vec{v}|\rho)$ distributions, one for each value of local density $\rho$. Once we get the distributions of speed reduction, we can generate random values $IP_i$ for speed reduction (IP represents the factors that impacts individuals as a function of present population), according to the local density of individual $i$ in video $v$. This reduction is incorporated to adjust our estimation method, given by:
\begin{equation}
\vec{X}^3_{t_e,i} = \vec{X}^2_{t_e,i} - IP_i.\vec{m}_{t,i}.(t_e - t).
\label{eq:ECTM}
\end{equation}
Equation~\ref{eq:ECTM} estimates position for agent $i$ in estimated time $t_e$, considering Physics and also that people in the video will keep their tracked motion vectors being penalized by the $EC_v$, which computes the environment complexity, and Weibull distribution, which generates the impact of population (factor $IP$) in the individuals estimation. In next section we show the changes proposed to estimate the future positions of people when the crowd flow is impacted by events that can change the expected flow.

\subsection{Third Step: Considering Crowd Flow Changes}

Here we classify the events to be dealt with in two types, as proposed previously by the authors~\cite{Bianco2017}, however adapting and testing that in video sequences. At first, the possibility of changing the environment by adding, changing or removing obstacles. Secondly, to increase or decrease the number of agents in the video. So, an event $k$  is defined by $e_k =  \{ t_k, \Delta t_k, \tau_k, \vec{O_k}, \vec{A_k} \}$, where $t_k$ is the event start time, $\Delta t_k$ is the event duration (can be unknown), $\tau_k$ defines the event type (obstacles $\tau_k$=1, population $\tau_k=2$ or both $\tau_k=3$), $\vec{O_k}$ and $\vec{A_k}$ describe obstacles and agents information. $\vec{O_k}$ is defined as $\{o_k, \vec{p_k}\}$, where $o_k$ is the number of obstacles, and $\vec{p_k}$ is the geometry for each obstacle (list of vertices, assuming a polygonal region, as defined by the user and mentioned in Section~\ref{sec:tracking}). $\vec{A_k}$ is defined here as $\{a_k, \vec{p_k}, \vec{v_k} \}$, where $a_k$ is the number of agents, $\vec{p_k}$ and $\vec{v_k}$ are positions and motion vectors for agents, respectively. Important to emphasize that are two possible ways to inform the events to be used in the estimation. Firstly, procedurally, by defining fictitious events that one wants to study the impact. Another way is to include some images (a short movie) that has to be tracked (as in Section ~\ref{sec:tracking}) in the same environment, and which captured information serve to parametrize an event. In the results discussed in Section~\ref{sec:results} we utilized the second option.

\begin{figure*}[h]
\centering
\includegraphics[width=16cm]{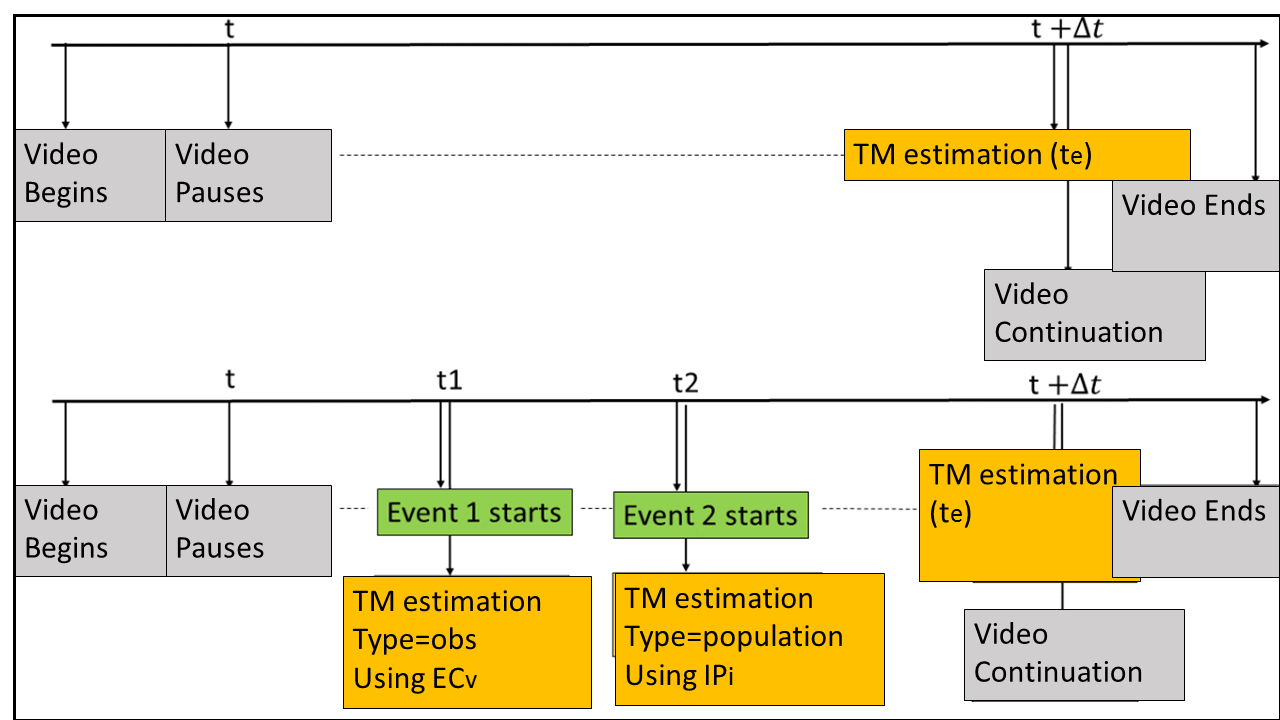}
\caption{Diagram showing the events in time line and the relationship with FF method execution.}
\label{fig::timeline}
\end{figure*}

Hence, we are interested in events that are triggered during the period of time when behaviors are going to be predicted. 
Let us consider a single event $e_k$, starting at frame $t_k$. In this case, the video stops at frame $t$ and one wants to estimate the positions of all individuals with 2 seconds in advance, i.e. $t_e = t + 60$, considering $30frames/second$. The initial estimation for each agent $i$ is computed using Physics, more precisely, it is given by Equation~\ref{eq:PDR1}.
Then, we used the two penalty functions defined in Equations ~\ref{eq:ECPos}
and ~\ref{eq:ECTM}, depending on the event type. An event $e_k$ with $\tau_k=1$ (type obstacle) should impact the environment complexity to be considered (Equation ~\ref{eq:ECPos}), while more people entering in the scene will impact using Equation ~\ref{eq:ECTM}. 
Figure~\ref{fig::timeline} illustrates one theoretical example to show the execution of our method. A video pauses (or stops) at frame $t$. The user asks to estimate the future positions of all individuals (present at frame $t$) at frame $t_e$. The timeline at the top of Figure shows an estimation without any triggered event, i.e. Equations 2, 3 and 4 are used to compute future positions of each individual at frame $t_e$, i.i $t+\Delta t$. On the bottom, we consider the same situation, however at times $t1$ and $t2$ two events happen, specifically of types $\tau_1=1$ (obstacles) and $\tau_2=2$ (population). In this case, our method estimates the future positions of people from $t$ to $t1$ using Equations 2, 3 and 4 however Equation 3 has values updated using Event 1 information. Then, estimations are made from frame $t1$ to $t2$ and in this case Equation 4 is updated with parameters from Event 2. Finally, estimation from time $t2$ to $t_e$ is performed, considering updated the event parameters, if they are still active.

\section{Experimental Results}
\label{sec:results}

\begin{figure*}[htb]
\begin{center}
\includegraphics[width = 17cm]{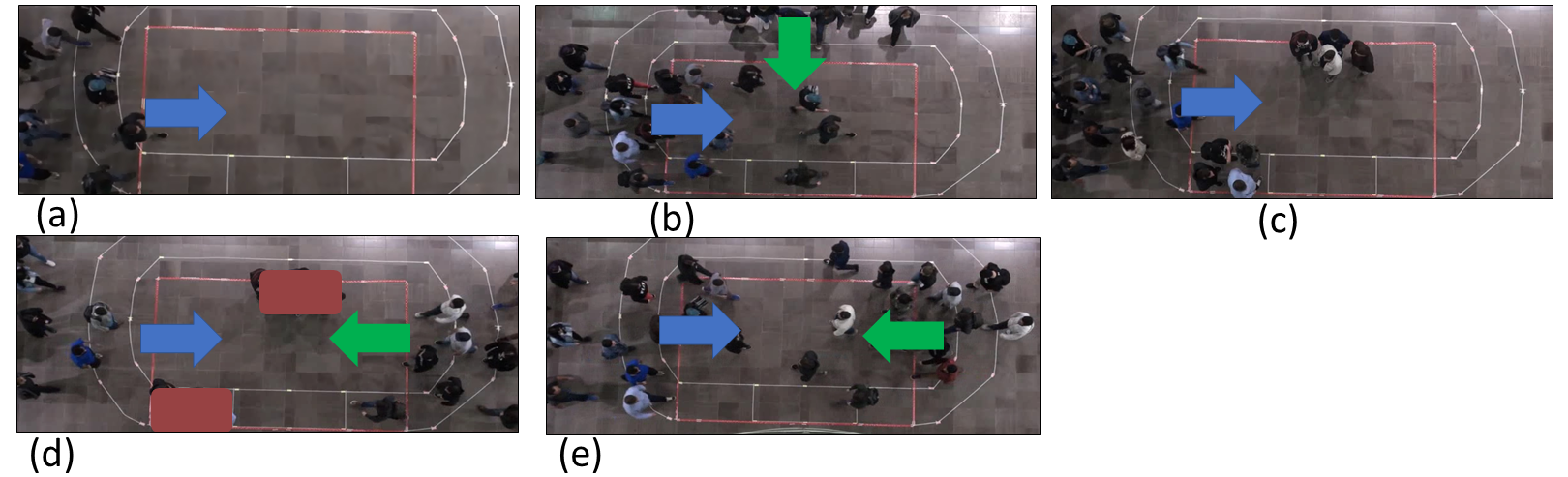}
\caption{Five scenarios filmed in the same environment - static camera. Flow directions of people is showed according to the arrows and in (e) two groups are formed and highlighted as 2 rectangles.}
\label{fig:cena}
\end{center}
\end{figure*}

In order to evaluate our work, we filmed 5 different scenarios in the same environment ($32m^2$) as illustrated in Figure~\ref{fig:cena}. In (a) there are 12 people walking from left to right and no event; 
In (b) there are 20 people also from left to right and six people from the top to the bottom of the image, however at frame 90, five people entry in the scene from the top to the bottom of the image (event $\tau=2$). In (c) and (e) two groups of people are formed to serve as obstacles (each one with dimensions $1x2$).
 In (c) there are 16 people and (at time 130) we considered the event (event $\tau=2$) of including more eight people. In (d), 11 people walk from left to right, 14 from right to left and we added more seven people (from left to right) distributed in two events $\tau=2$ (according to Table~\ref{tab::error}), while in (e) 12 walk from each side in the image having two obstacles representing the event $\tau=1$. 

\begin{figure*}[htb]
\begin{center}
\includegraphics[width = 16cm]{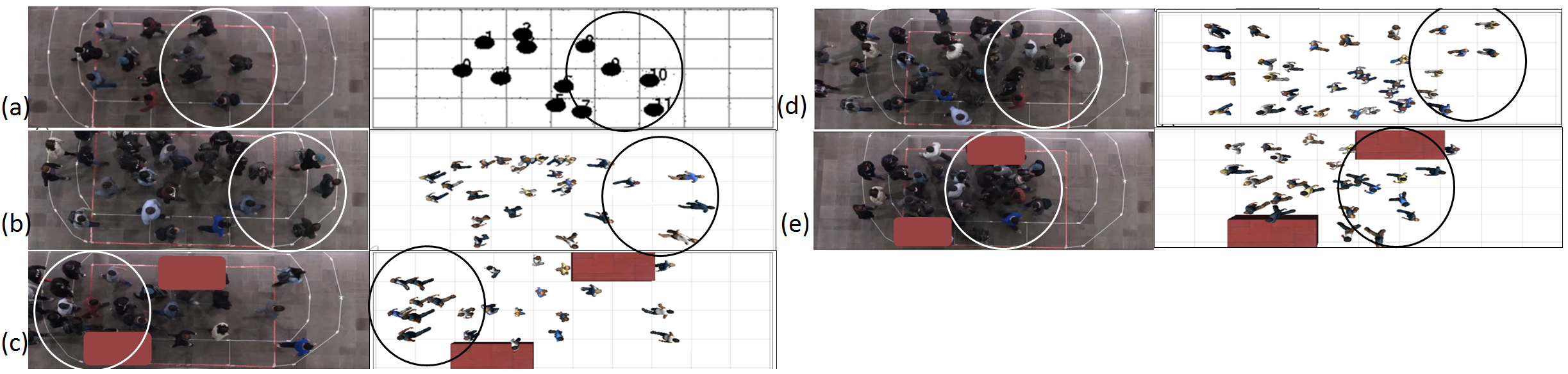}
\caption{Five tested videos and respectively simulation results (with white background).}
\label{fig:resultsSo}
\end{center}
\end{figure*}

In the first, fourth and fifth scenario we stopped the videos at frame 60 and estimate the positions of all individuals at frame 130, in the second scenario we stopped the videos at frame 80 and estimate the positions at frame 130 and third scenario we stopped the videos at frame 70 and estimate the positions at frame 140, as described in Table~\ref{tab::error}. We measured the average error of estimated position for individuals using Equation~\ref{eq:Error1}. 
Quantitative results can be seen in Table~\ref{tab::error}, while qualitative assessment of results can be observed in Figure~\ref{fig:resultsSo}, when compared real video and simulations.
Since scenarios can be advanced in different number of frames, we computed the error per frame. As could be observed in the table, the presented errors values are small (maximum achieved average error is $2.72cm/frames$ observed in scenario (d)) which indicate that our method accomplish adequately with the goal to provide a method to estimate future positions of people in video sequences in next seconds. The worst value was achieved in scenario (d) and it is explained because this video had two events to treat in the method (frames 70 and 90) that includes more people to be considered.  
In addition, new people (7 individuals) entry in the scene at the right moment of crossing between the two groups (11 from left to right and 14 from right to left),
generating two events (frames 70 and 90). As one can remark, the situation shown in scenario (d) impacts more the obtained error value than the situation in scenario (e), which has two flows of people and the event is only the obstacle.


\begin{table}[h]
\centering
\caption{Results of individuals estimation at each scenario. In the first column we can see the specification of scenario to be discussed and $[t,t_e]$ that represent the time where the video ends and estimation time, respectively. In the second column we see the happened events and their information for all scenarios except for the first one, where no event happened. In the last column we present the average error observed in $cm/frame$.}
\label{tab::error}
\begin{tabular}{|c|c|c|}
\hline
\small
Id Scenario  & Event & Avg Dist Error \\
as Fig.~\ref{fig:cena} $[t,t_e]$ & information  & ($cm/frame$) \\ \hline
(a) [60,130]  & No event & 0.8  \\ \hline
(b) [80,130] & 1 Events $\tau=2$ &     \\ 
& $t_1=90$, $a_1=5$  &   1.92  \\ 
\hline
%
(c) [70,140] & 1 Event $\tau=2$ &      \\ 
& $t_1=130$, $a_1=8$  & 1.36    \\ 

\hline
(d) [60,130] &  2 Events $\tau=2$  &     \\
& $t_1=70$, $a_1=4$  &  2.72   \\ 
& $t_2=90$, $a_2=3$  &     \\ 
\hline
(e) [60,130] &  1 Event $\tau=1$   &   \\ 
  &  $t_1=70$, $o_1=2$  & 1.6    \\ 
\hline
\end{tabular}
\end{table}

\section{Final Considerations}
\label{sec:final}

This paper presents an extension of~\cite{Bianco2016} model to estimate crowd motion in a future time (fast forwarding), while treating events~\cite{Bianco2017}, specifically for real video sequences. The maximum error observed in the 5 studied cases is $2.72$cm/frame when two events of inclusion of people occurred in the same estimation process. In addition, as illustrated in Figure~\ref{fig:resultsSo}, one can observe that the qualitative results present very similar positions of people in those scenarios when compared to the real video sequences. 
Future work should include more tests with other events and videos containing more people.


\bibliographystyle{splncs03}  

\bibliography{biblio.bib}  

\end{document}